# TV-SAM: Increasing Zero-Shot Segmentation Performance on Multimodal Medical Images Using GPT-4 Generated Descriptive Prompts Without Human Annotation

Zekun Jiang[#], Dongjie Cheng[#], Ziyuan Qin, Jun Gao, Qicheng Lao, Abdullaev Bakhrom Ismoilovich, Urazboev Gayrat, Yuldashov Elyorbek, Bekchanov Habibullo, Defu Tang, LinJing Wei, Kang Li[*], Le Zhang[*]

**Abstract:** This study presents a novel multimodal medical image zero-shot segmentation algorithm named the text-visual-prompt segment anything model (TV-SAM) without any manual annotations. The TV-SAM incorporates and integrates the large language model GPT-4, the vision language model GLIP, and the SAM to autonomously generate descriptive text prompts and visual bounding box prompts from medical images, thereby enhancing the SAM's capability for zero-shot segmentation. Comprehensive evaluations are implemented on seven public datasets encompassing eight imaging modalities to demonstrate that TV-SAM can effectively segment unseen targets across various modalities without additional training. TV-SAM significantly outperforms SAM AUTO (P<0.01) and GSAM (P<0.05), closely matching the performance of SAM BBOX with gold standard bounding box prompts (P=0.07), and surpasses the state-of-the-art methods on specific datasets such as ISIC (0.853 versus 0.802) and WBC (0.968 versus 0.883). The study indicates that TV-SAM serves as an effective multimodal medical image zero-shot segmentation algorithm, highlighting the significant contribution of GPT-4 to zero-shot segmentation. By integrating foundational models such as GPT-4, GLIP, and SAM, the ability to address complex problems in specialized domains can be enhanced.

- Zekun Jiang, Jun Gao, and Le Zhang are with the College of Computer Science, Sichuan University, Chengdu, Sichuan, China. E-mail: jiangzekun@stu.scu.edu.cn; gaojun@stu.scu.edu.cn; zhangle06@scu.edu.cn.
- Dongjie Cheng and Kang Li are with the West China Biomedical Big Data Center, West China Hospital, Sichuan University, Chengdu, China. E-mail: chengdongjie@stu.scu.edu.cn; likang@wchscu.cn.
- Ziyuan Qin is with Case Western Reserve University, Cleveland, US.
- Qicheng Lao is with School of Artificial Intelligence, Beijing University of Posts and Telecommunications, Beijing, China.
- Abdullaev Bakhrom Ismoilovich, Urazboev Gayrat, Yuldashov Elyorbek, and Bekchanov Habibullo are with the Urgench State University, Urgench, Uzbekistan.
- Defu Tang is with College of Animal Science and Technology, Gansu Agricultural University, Lanzhou, China.
- LinJing Wei is with College of Information Science and Technology, Gansu Agricultural University, Lanzhou, China.

\# The authors contributed equally to this work.
* To whom correspondence should be addressed.

## 1 Introduction

As the first vision foundation model, the segment anything model (SAM) emergence has spurred researchers worldwide to explore its zero-shot segmentation performance in various scenarios [1-3], especially in the field of medical imaging [4-14]. In clinical practice, medical images are the primary basis for decision-making in disease diagnosis and treatment. Since medical imaging typically consists of multiple modalities with different organs of the human body, the segmentation scenarios are usually very complex [15]. Therefore, if we can find an effective zero-shot segmentation method that can eliminate the need to train related segmentation models via supervised, weakly supervised, or transfer learning, it could help us save considerable data and computational resources.

Currently, SAM has three prompt working modes, which are no-prompts, point prompts, and box prompts [7], but it cannot directly use text prompts for auto segmentation. The open-source community has provided several alternative solutions, such as GroundedSAM [16], but they still rely on human annotations as prompt inputs. Thus, we consider that these methods are insufficient for automatic medical image segmentation. Furthermore, the current evaluation studies of SAM zero-shot segmentation mostly rely on points or boxes generated from ground truth masks as prompt inputs [6, 9]. Undoubtedly, these methods are not applicable for dataset or domains lacking annotated information, and thus cannot be directly generalized for use in clinical applications.

Therefore, exploring a zero-shot segmentation algorithm for multimodal medical images is a worthwhile endeavor. In previous studies, Qing et al. [17] proposed the use of a vision language model (VLM) to obtain bounding boxes (bboxes), which could facilitate segmentation. Since accurately providing text prompts for VLM is challenging, and the experimental results indicate that the representations learned by existing VLMs still exhibit a significant domain gap with medical data [17], relying solely on VLMs fails to yield sufficiently high-quality results.

Recently, the large language model (LLM), with its unique capabilities such as emergence and chain of thought [18, 19], has brought the foundation model concept into the public eye. In particular, generative pre-trained transformer 4 (GPT-4) has essentially become the most powerful intelligence model to date [20]. Thus, it could be a completely viable approach for us to employ GPT-4 as a knowledge source to design a better expressive prompt for medical concepts.

For these reasons, this study developed a novel zero-shot segmentation algorithm by integrating GPT-4 prompts, VLM, and SAM. We extensively evaluated the approach on multimodal medical image datasets, and compared it with other zero-shot segmentation methods to demonstrate its superiority.

## 1.1 Related work

**Vision language model (VLM):** Zero-shot segmentation enables models to recognize and segment objects in downstream tasks that have not been seen before without additional training. A groundbreaking work concerning zero-shot learning in computer vision is contrastive language-image pre-training (CLIP) [21]. As a VLM, CLIP leverages contrastive learning on images and text to generate a shared multimodal representation. Based on this multimodal representation, semantic segmentation of unseen images can be achieved. Consequently, CLIP and its subsequent extensions have been applied to medical image segmentation tasks. Grounded language-image pre-training (GLIP) is a more advanced VLM that extends CLIP's learned representations at the image level to object-level representations [22]. Therefore, it is better suited for phrase and image object localization. Liu et al. [23] has built a CLIP-driven universal model for organ segmentation and tumor detection. Qin et al. [17] also demonstrated that rapid adaptation to unknown medical image data for target detection can be achieved through the integration of prompts and GLIP.

Although preliminary exploratory efforts have already made great progress, how to optimize the prompts given to VLMs and automatically extract expressive high-quality prompts from medical images is still an important topic that needs further exploration.

**Segment anything model (SAM):** With the emergence and development of SAM, zero-shot learning has become increasingly popular. Generalist foundation models learn more diverse and extensive multimodal representations and possess unprecedented capabilities. Since the publication of the SAM, research teams from around the world have extensively assessed the segmentation performance of the SAM in diverse medical imaging scenarios from various perspectives. Several previous studies collected dozens of medical image datasets to assess SAM's segmentation performance [6, 9], but no particularly intriguing conclusions were drawn. In summary, the SAM without any prompts often results in poor segmentation performance, whereas it shows very stable performance with multiple points or bbox prompts. This problem could be caused by the large domain gap between medical images and natural images.

For this reason, further studies have fine-tuned the SAM and developed specialized models for medical image processing, such as the MedSAM [5], and SAM-Med2D/3D [24, 25].

On the one hand, these methods all require points or bboxes as visual prompts and visual prompts need manual input, failing to achieve satisfactory segmentation in zero-shot settings without any prompts. Additionally, works such as SAM-Med2D/3D request a substantial data and computational resource investment during the fine-tuning process [24, 25]. Therefore, the exploration of the SAM application is still ongoing.

On the other hand, this leads to a debate regarding generalist and specialist models. We often argue that for medical imaging, specialist models, once fine-tuned, may prove more effective. However, previous studies [26] have shown that generalist models still possess untapped potential with ingenious prompt engineering. Even for the concepts not present in their pre-training, these models can be effectively 'activated' by prompts that employ analogical understanding of the concepts along with detailed descriptions of the visual nuances [26, 27]. Consequently, we can achieve performances unattainable by a single generalist model alone by blending various generalist models.

**Large language model (LLM):** LLM, especially GPT-4, being the most powerful knowledge source to date, are fully capable of designing higher quality prompts for VLM and SAM.GPT-4 has been explored for medical image processing [20, 28]. However, using GPT-4 directly for image processing involves various challenges; mainly because when facing medical tasks, GPT-4 becomes particularly cautious. GPT-4 is not a medical professional (e.g., radiologist or pathologist), making it unable to provide more precise answers [28]. In several cases, it may even refuse to recognize certain medical images. Therefore, although the current version of the GPT-4 is not recommended for direct use in real-world clinical diagnosis [28], the GPT-4 can be indirectly employed by generating rich descriptive prompts, thereby activating and enhancing the recognition and segmentation performance of the VLM and SAM.

In summary, VLMs, SAM, or GPT-4 have shown promising potential in zero-shot segmentation, but are still not perfect. Currently, there is no research exploring the effective integration of these methods to achieve better segmentation results.

## 1.2 Key contributions

Therefore, this study introduces a novel zero-shot segmentation algorithm for medical images, named text-visual-prompt SAM (TV-SAM), which is co-driven by GPT-4 prompts, VLM (GLIP), and SAM. The TV-SAM contributes the following:

1. We propose a GPT-4-driven SAM zero-shot segmentation algorithm for medical images and conduct comprehensive evaluations for multimodal medical image datasets with eight modalities and nine targets.

2. We developed a novel zero-shot segmentation algorithm framework for medical images. By integrating the GPT-4, GLIP, and SAM, we not only automatically generate text prompts and visual prompts based on input images, but also segment the objects without any additional human prompts.

3. Our approach demonstrates better segmentation performance than the SAM without prompts and the GSAM. In several scenarios, the performance approximates the SAM with gold standard bbox prompts. The experimental results indicated that while this method can function as a zero-shot segmentation tool, the overall segmentation performance still needs improvement, particularly for radiological images.

## 2 Methodology

Before delving into the specifics, we first provide an overview of our workflow from a high-level perspective in Fig. 1. In general, our TV-SAM methodology is a multi-stage prompt learning algorithm framework that leverages the prompt information between different foundational models. It consists of three primary stages: first, the generation of text prompts; second, the creation of visual prompts; and third, the mask decoding stage.

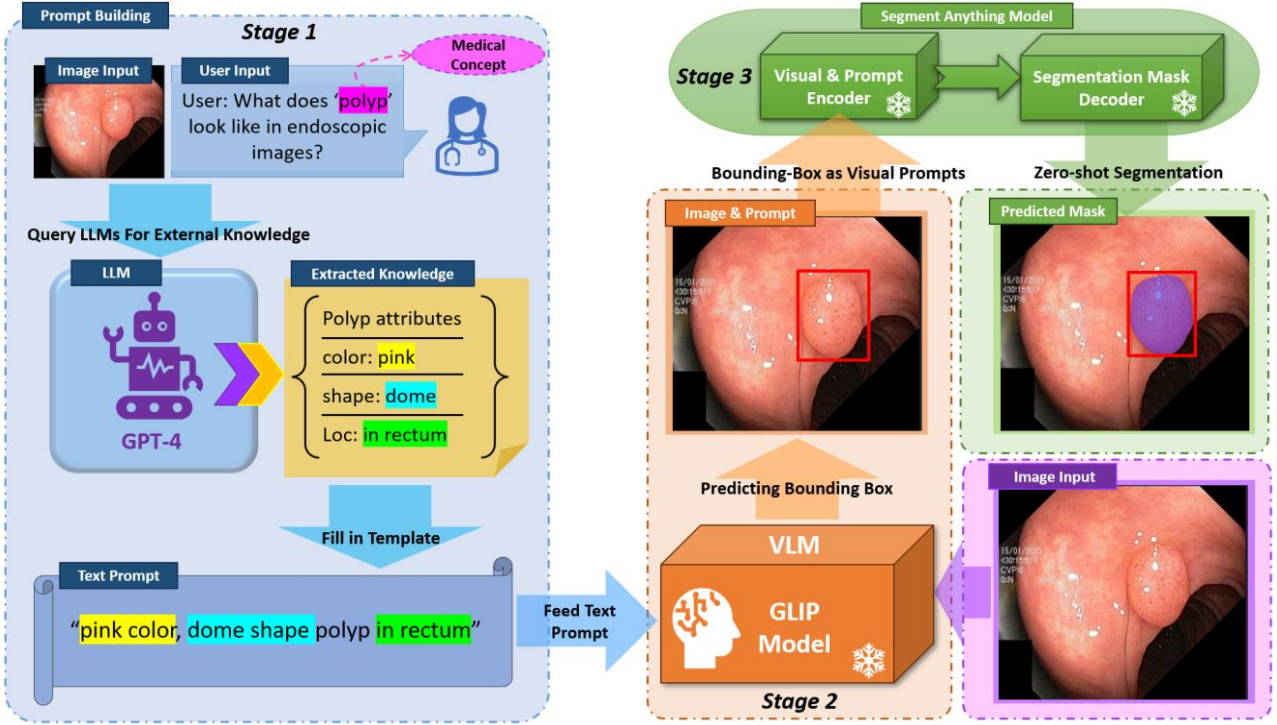

**Fig. 1 Proposed zero-shot algorithm framework for medical image segmentation**

In the first stage, we employ GPT-4 as a knowledge source to generate text prompts that vividly describe medical concepts with the expressive detail of medical images.

In the second stage, we employ pre-trained VLMs to identify the probable regions of the medical concepts based on the provided prompts, typically via a bounding box format.

In the third stage, the bounding boxes for the areas of interest are employed as visual prompts to assist the SAM in more accurately predicting the segmentation masks for these areas.

In summary, users only supply unlabeled medical images along with the concepts, objects, or abnormalities of interest. Subsequently, TV-SAM autonomously executes the three stages and generates the corresponding segmentation masks.

## 2.1 GPT-4 as knowledge sources for expressive prompts design

To generate the text prompts, we present the GPT-4 driven expressive prompts design algorithm as Stage 1 of Fig. 1. Here, we employ GPT-4 as an external knowledge sources to generate detailed text prompts for medical concepts in medical images.

$$F_I = ImageEncoder(I) \quad (1)$$

$$F_P = PromptEncoder(P) \quad (2)$$

$$F_{fusion} = CrossAttention(F_I, F_P) \quad (3)$$

Specifically, we feed images $I$ and a dialog template as $P$ into the GPT-4 model. GPT-4 employs a separate vision encoder to extract high-level features $F_I$ from image $I$ via Eq. (1) and uses a text encoder to obtain $F_P$ from $P$ via Eq. (2). Through cross-attention, the representations from both are fused to obtain $F_{fusion}$ and are embedded into the LLM via Eq. (3).

$$D_{LLM} = Decoder(F_{fusion}) \quad (4)$$

Then, a decoder returns the specific information $D_{LLM}$ about the target concept via Eq. (4), including attributes such as color, shape, and location.

Upon acquiring these attributes, we employ an additional template, Temp_prompt, to construct the final text prompts by incorporating the obtained attributes. This process can be executed independently or directly embedded into the dialog template, guaranteeing that the information returned by the GPT conforms to the final prompt format.

## 2.2    Generating visual prompts via VLM

As previously articulated, the application of the SAM in medical imaging predominantly relies on human-labeled visual prompts, such as bounding boxes or center points. While we acknowledge that providing visual prompts demands less effort than creating ground truth segmentation masks, the curation of visual prompts for all the samples remains a nontrivial task. This is primarily due to the requirement of expert-level knowledge to identify areas of interest accurately. For these reasons, we employ VLMs to generate visual prompts automatically as Stage 2 of Fig. 1.

Here, we select GLIP [22] as the foundational VLM, the input image $I$ is fed back into the GLIP image encoder to obtain $F_I$, as shown in Eq. (1), and the previously obtained descriptive prompts $D_{LLM}$ serve as text prompts $P$ into the GLIP prompt encoder to obtain $F_P$ via Eq. (2).

$$B_{VLM} = BboxDecoder(F_{fusion}) \tag{5}$$

Then, $F_{fusion}$ is generated via Eq. (3) and embedded into a GLIP bbox decoder to predict bounding boxes $B_{VLM}$ via Eq. (5), this essentially achieves zero-shot object detection.

Importantly, the quantity of predicted bounding boxes typically exceeds what is necessary, and thus requires reduction via the non-maximum suppression algorithm [29]. Moreover, we further refine the selection by filtering out any predicted boxes that exhibit a confidence score below a predetermined threshold.

## 2.3    SAM zero-shot segmentation with visual prompts

The SAM can accept three types of prompts: point, box, and text, but it should be noted that the text prompt variant is not yet publicly available. Previous research [9] has indicated that the box prompt is particularly effective when the SAM is applied in the medical domain, and that the box mode of the SAM achieves optimal performance across a range of medical image datasets for our studies [7]. Therefore, we implement a box prompts-based SAM zero-shot segmentation algorithm as Stage 3 of Fig. 1.

$$M_{SAM} = MaskDecoder(F_{fusion}) \tag{6}$$

As shown in Eq. (1-3), we reinput the image $I$ into the SAM image encoder to obtain $F_I$, and input the bounding boxes $B_{VLM}$ predicted by GLIP as visual prompts $P$ into the SAM prompt encoder to obtain $F_P$. Then, $F_{fusion}$ is input into the SAM mask decoder to generate accurate segmentation masks $M_{SAM}$ via Eq. (6).

## 3  Experiment

### 3.1    Datasets

For a comprehensive evaluation of the efficacy of our approach in clinical practice, we collected seven publicly available datasets as multimodal medical imaging datasets, encompassing eight distinct medical imaging modalities and nine segmentation objects. Table 1 lists the general distribution of the datasets. Because our algorithm does not need any training, we directly use the test set of the public datasets as the evaluation data. Detailed descriptions of the datasets are provided below:

**(1)  Polyp benchmark:** The diagnosis of gastrointestinal diseases often involves endoscopic imaging. This dataset consists of endoscopic images specifically focused on polyps and the associated

Table 1. Overall distribution of the datasets.

| Dataset | Modality | Segmentation object | Num. evaluated data |
|---|---|---|---|
| Polyp benchmark [30] | Endoscope | Polyp | 602 |
| ISIC 2018 [31, 32] | Dermoscopy | Melanoma | 1002 |
| WBC [33] | Microscopy | Cell | 300 |
| BUSI [34] | Ultrasound | Breast cancer | 1312 |
| TN3K [35] | Ultrasound | Thyroid nodules | 614 |
| COVID-19 Database [36, 37] | X-ray | Lung | 10192 |
| CHAOS [38] | CT, T1 MRI, T2 MRI | Liver, Spleen, Kidney | 300, 319, 302 |

neoplastic changes in the colon and rectum, with the corresponding masks delineating the polyps.

**(2) ISIC 2018:** Skin diseases are typically preliminarily diagnosed through regular photographic images. This dataset comprises a substantial number of skin disease images, involving conditions such as melanoma and basal cell carcinoma, with segmented masks for each lesion.

**(3) WBC dataset:** The diagnosis of blood disorders requires microscopic imaging of blood cells. This dataset comprises hundreds of images of blood cell morphology along with corresponding masks delineating the cell nuclei.

**(4) BUSI dataset:** Ultrasound scanning is a commonly used diagnostic tool for detecting breast diseases. This dataset comprises ultrasound images of 600 female patients aged between 25 and 75 years old, with segmentation masks delineating the tumor regions.

**(5) TN3K dataset:** The diagnosis of thyroid diseases relies primarily on ultrasound examinations. This dataset comprises over three thousand thyroid ultrasound images, with masks delineating thyroid nodules.

**(6) COVID-19 Database:** The diagnosis of lung diseases is inseparable from X-rays and chest CT scans, and this is a database related to COVID-19 chest X-ray images and lung masks.

**(7) CHAOS dataset:** The diagnosis of thoracic and abdominal diseases typically involves radiological imaging, primarily CT and MRI. This dataset combines CT-MRI healthy abdominal organ segmentation, including CT data with liver masks, as well as T1 and T2 MRI data with segmentation masks for the liver, spleen, and kidneys.

### 3.2    Comparative experiment

To demonstrate the superiority of the TV-SAM, we conduct comparative evaluations with various SAM benchmark algorithms, including the SAM with no-prompt (SAM AUTO), the SAM with gold standard bbox prompts (SAM BBOX), and only using the GLIP and SAM (GSAM).

**SAM AUTO:** This is also called the SAM 'everything mode', where images are input into the SAM without any prompts, and then the SAM generates segmentation masks for targets in the image directly based on the image decoder. However, because there are no prompts for specific targets, the performance of this method is usually not very good, although it can demonstrate better segmentation performance in image scenes with a single clear target.

**SAM BBOX:** This is implemented based on the SAM 'bbox mode'. Among the three modes of the SAM, the 'bbox mode' is known to have the best performance. Therefore, the gold standard boxes from the public datasets are used as input prompts, which are input into the SAM along with the images. Through the prompt encoder, image encoder, and mask decoder, we can obtain accurate segmentation results.

**GSAM:** To demonstrate the contribution of LLM to our study, we also implemented the GSAM algorithm. This algorithm uses only GLIP and SAM to perform segmentation, directly inputting the image and the name of the object to be segmented into GLIP.

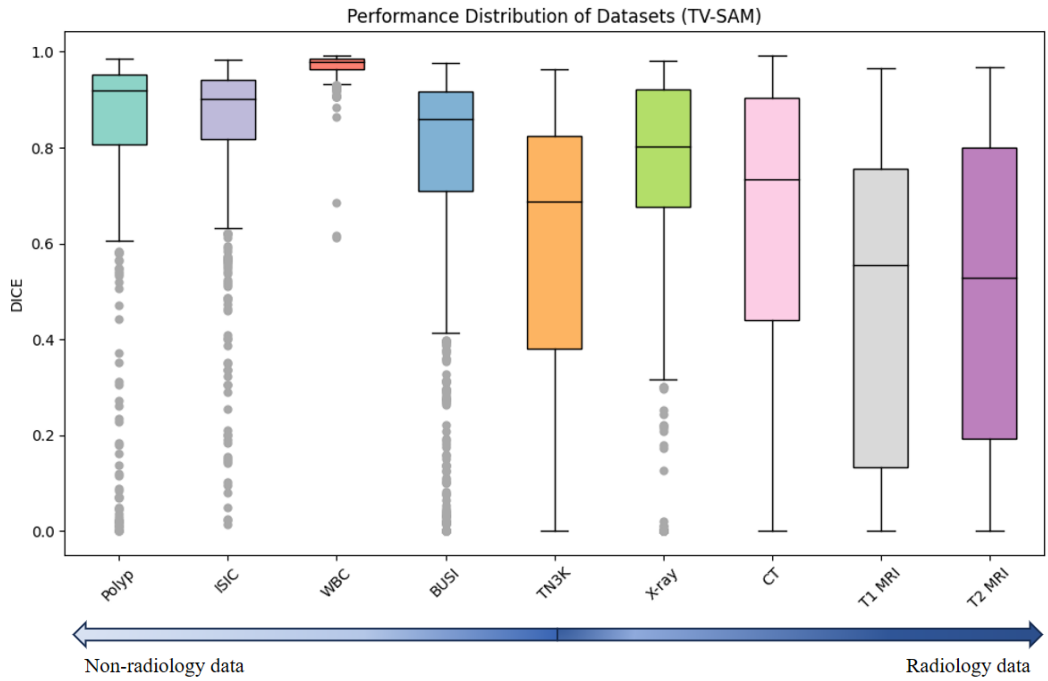

**Fig. 2 Performance distribution of TV-SAM in multimodal datasets**

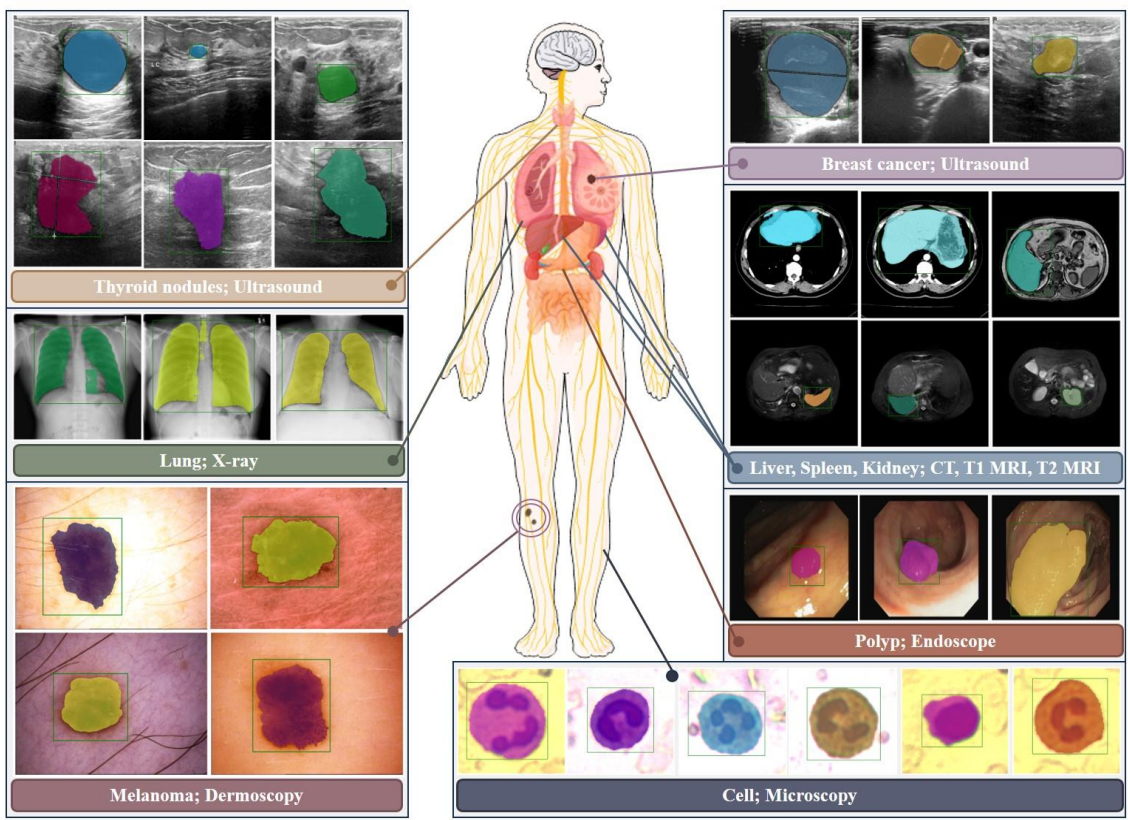

**Fig. 3 Overview of our zero-shot segmentation performance across multiple organs in multimodal medical imaging**

GLIP identifies targets in the image based solely on the prompts of the object name and returns bbox prompts to the SAM for further segmentation.

### 3.3 Statistical analysis

The segmentation results are evaluated by the Dice coefficient, as shown in Eq. (7).

**Table 2. Zero-shot segmentation performance comparison.** The value is the average Dice coefficient, with the bold fonts indicating the best performance.

| Methods | Polyp | ISIC | WBC | BUSI | TN3K | X-ray | CT | T1 MRI | T2 MRI |
|---|---|---|---|---|---|---|---|---|---|
| SAM BBOX | 0.909 | 0.8833 | 0.97 | 0.8421 | 0.78 | 0.7813 | 0.829 | 0.5539 | 0.605 |
| TV-SAM | **0.831** | **0.853** | **0.968** | **0.751** | **0.588** | **0.788** | **0.636** | **0.47** | **0.488** |
| GSAM | 0.453 | 0.777 | 0.965 | 0.734 | 0.412 | 0.6297 | 0.591 | 0.47 | 0.441 |
| SAM AUTO | 0.654 | 0 | 0.955 | 0.5312 | 0.4324 | 0.565 | 0.63 | 0.4323 | 0.406 |

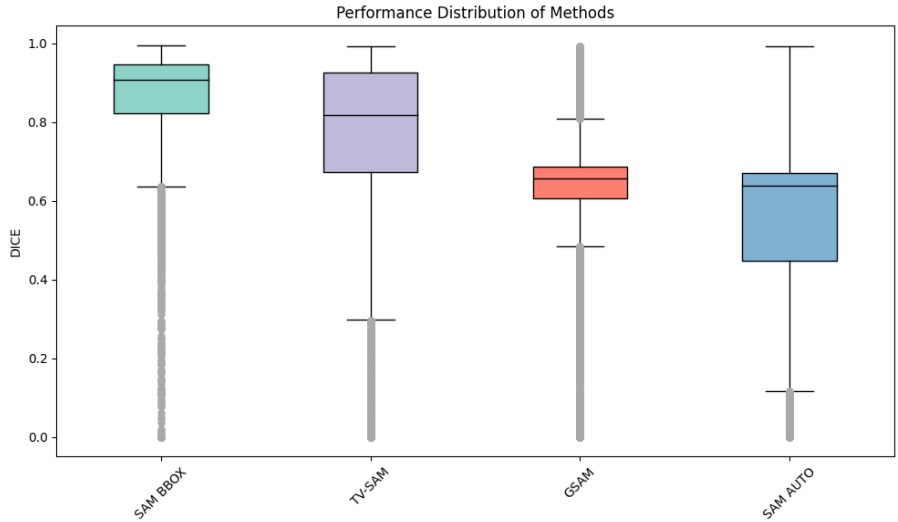

**Fig. 4 Performance distributions of different zero-shot segmentation methods**

$$Dice(M, M^*) = \frac{2|M \cap M^*|}{|M| + |M^*|} \quad (7)$$

Here, $M$ represents the actual segmentation mask, whereas $M^*$ represents the predicted segmentation mask. The related distributions are represented by mean values and 95% CIs. Statistical tests between different segmentation performances employ the t-test [39-44], with P<0.05 indicating a significant difference.

## 4 Results and Discussion

This section presents the results from three key aspects: (1) the quantitative evaluation of our algorithm (TV-SAM) in multimodal medical imaging datasets, (2) the comparison of TV-SAM with three others commonly used SAM-based zero-shot algorithms and supervised learning algorithms, and (3) the comparison of segmentation performance under different bounding box selection strategies.

For the first aspect, Fig. 2 quantitatively assesses the segmentation performance of TV-SAM for each imaging modality dataset.

For non-radiological images, our method achieves an average Dice score above 0.8, with Dice coefficients of 0.831, 0.853, and 0.968 for the Polyp, ISIC, and WBC datasets, respectively.

**Table 3. TV-SAM versus SOTA on non-radiological datasets.** The value is the average Dice coefficient, and the bold values indicate the best performance.

| Methods | Polyp | ISIC | WBC |
|---|---|---|---|
| TV-SAM | 0.831 | **0.853** | **0.968** |
| SOTA | **0.898** | 0.802 | 0.883 |

However, for radiological images, the segmentation performance decreases. For the BUSI and COVID-19 X-ray datasets, the Dice scores are close to 0.8, being 0.751 and 0.788 respectively, while on the other datasets such as CT and MRI, the metrics are less than 0.7.

Moreover, Fig. 3 shows several accurately predicted zero-shot segmentation cases, which help us observe the segmentation results of TV-SAM more visually on cross-organ and cross-modal medical imaging data. Importantly, TV-SAM can directly segment unseen multimodal medical image targets without an additional training process, which provides a new strategy and evidence to solve complex issues with limited samples and multimodality in the medical imaging field.

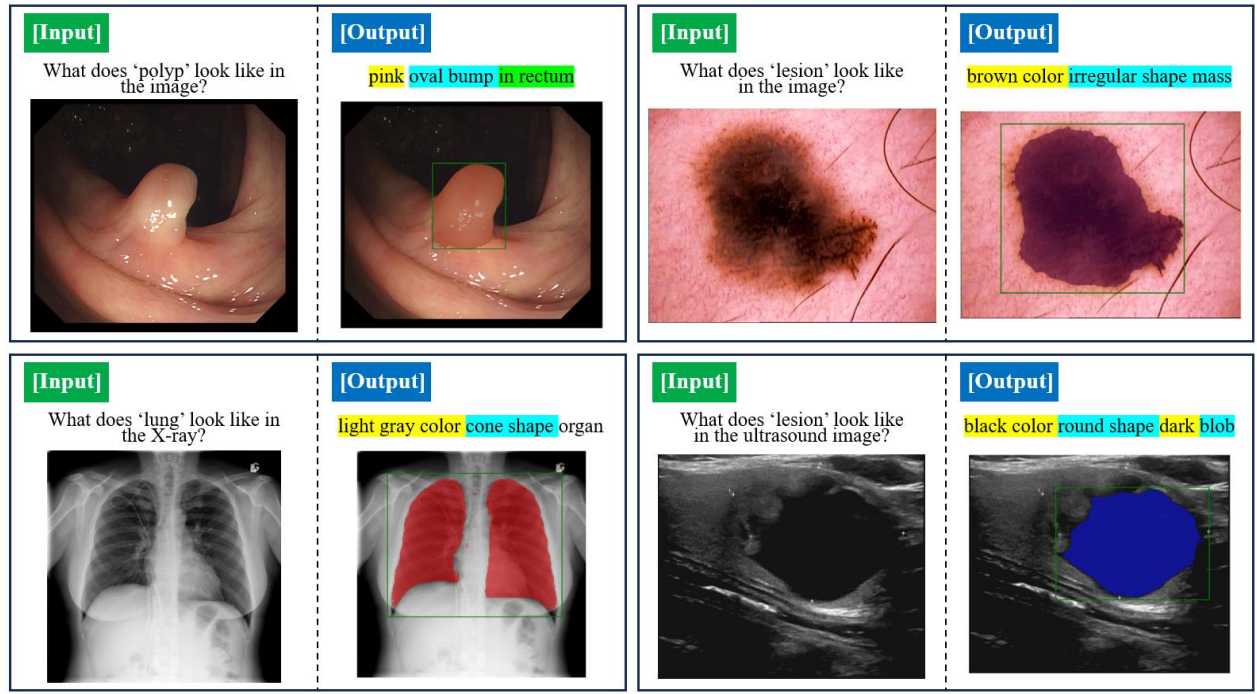

**Fig. 5 Clinical implementation cases of the TV-SAM algorithm. The yellow area of the descriptive text represents color information, the blue area represents shape information, and the green area represents location information.**

In other words, fully leveraging the general knowledge representations of foundational models via GPT-4 and SAM, along with appropriate prompt techniques for knowledge activation, can be directly applied to address complex problems in specialized domains.

For the second aspect, Table 2 provides a performance comparison of TV-SAM with SAM AUTO, SAM BBOX, and GSAM across different datasets, which demonstrates the advantages of TV-SAM as a novel zero-shot segmentation algorithm. Additionally, Fig. 4 shows a comparison of the segmentation performance distributions of these methods across the entire multimodal medical imaging dataset.

Table 2 and Fig. 4 show that, on the one hand, different zero-shot methods generally exhibit the same pattern across various modality datasets, that is, greater average segmentation performance on non-radiological images, and less performance on radiological images such as CT and MRI; On the other hand, TV-SAM is better than SAM AUTO ($P<0.01$) and GSAM ($P<0.05$) in overall performance without any additional prompts; moreover, not only is the performance of the TV-SAM method almost the same as the SAM BBOX method, which relies on gold standard bbox prompts as input ($P=0.07$), but TV-SAM even outperforms the SAM BBOX method in terms of X-ray data (0.788 versus 0.7813).

Additionally, to further demonstrate the effectiveness of the TV-SAM, we select three non-radiological image datasets from Fig. 2 with the best segmentation performance for comparison, such as Polyp, ISIC, and WBC.

Our TV-SAM was compared against the state-of-the-art (SOTA) results for these datasets [30, 45, 46]. These SOTA outcomes were obtained through supervised learning on the training set and evaluated on the testing set. The classical deep learning segmentation algorithms they used include the parallel

10　　　　　　　　　　　　　　　　　　　　　　　　　　　　　　　　　　　　　　　　　　Big Data Mining and Analytics, 2024**Table 4. Performance comparison under different bounding box selection settings.** The values represent the average Dice coefficient and the values in bold fonts indicate the best performance.

| Methods | Polyp | ISIC | WBC | BUSI | TN3K | X-ray | CT | T1 MRI | T2 MRI |
|---|---|---|---|---|---|---|---|---|---|
| TOP-1 | 0.698 | 0.837 | 0.951 | 0.596 | 0.337 | 0.7 | 0.52 | 0.152 | 0.2 |
| TOP-2 | 0.739 | 0.848 | 0.965 | 0.669 | 0.439 | 0.751 | **0.636** | 0.249 | 0.33 |
| TOP-3 | 0.760 | 0.85 | **0.968** | 0.699 | 0.49 | 0.778 | 0.636 | 0.322 | 0.4 |
| TOP-5 | 0.792 | 0.852 | 0.968 | 0.73 | 0.556 | 0.786 | 0.636 | 0.401 | 0.444 |
| TOP-10 | **0.831** | **0.853** | 0.968 | **0.751** | **0.588** | **0.788** | 0.636 | **0.472** | **0.488** |

reverse attention segmentation network (PraNet) [30], an improved version of U-Net (called DCSAU-Net) [45], and a novel dual-task segmentation network [46]. As shown in Table 3, the performance of TV-SAM is comparable to that of SOTA in the Polyp benchmark and even exceeds that of SOTA in the ISIC and WBC datasets.

Furthermore, the significant performance improvement of TV-SAM over GSAM (Table 2) highlights the crucial contribution of GPT-4 in zero-shot segmentation. The rich descriptive prompts obtained from images by GPT-4 can further increase the accuracy and stability of the VLM and SAM.

In some datasets, such as ISIC and WBC, TV-SAM outperforms SOTA on unseen data without any manual prompts (Table 3), which further supports the paradigm shift brought about by foundation models in deep learning technology [47], emphasizing the growing importance of zero-shot and few-shot learning in practical applications.

In Fig. 5, we further provide clinical implementation cases of the TV-SAM to enhance the understanding of its algorithm and the credibility of its clinical application. By inputting medical images of any modality and asking questions about the segmentation target, TV-SAM can output text and visual prompts related to the target, as well as the final segmentation mask, through multi-stage prompt learning. Interestingly, the descriptive information regarding the color and shape of the target is sufficient for the VLM to recognize the target.

Importantly, the text and visual prompts in our method are autonomously generated and do not necessitate human intervention. This fully automated procedure operates independently of any human labeling or manual interfacing, even under zero-shot settings. Therefore, TV-SAM is more convenient and efficient than the conventional SAM methods and supervised learning methods.

Considering that the VLM can predict multiple bounding boxes, we further explore how to select bbox prompts to optimize SAM segmentation. As shown in Table 4, TOP-1 refers to using the highest confidence bbox predicted by VLM as the SAM prompt for segmentation, whereas TOP-k involves selecting the top k bboxes based on confidence scores and inputting them into the SAM simultaneously for optimal segmentation selection.

The findings suggest that while using the highest confidence bbox as the SAM prompt yields good segmentation performance, it is not the best choice.

Additionally, moderately increasing the number of alternative bboxes and employing the SAM to operate in parallel can improve segmentation performance to some extent, especially for datasets featuring multiple target concepts or objects within a single image, which provides some guidance on how the SAM selects bbox prompts.

## 5 Conclusion

We propose a novel multimodal medical image zero-shot segmentation algorithm, namely TV-SAM, which achieves efficient target segmentation by integrating LLM, VLM, and SAM. We perform a comprehensive evaluation on seven public datasets across eight medical imaging modalities, demonstrating the outstanding zero-shot segmentation performance of the TV-SAM.

Extensive experimental results demonstrate our TV-SAM from three aspects.

First, as a novel zero-shot segmentation algorithm, TV-SAM can effectively segment unseen targets across various modalities without extra training, highlighting the potential of a foundational model-driven prompt algorithm to solve the challenges of limited samples and multimodality in medical imaging.

Second, TV-SAM significantly outperforms SAM AUTO and GSAM, closely matches the performance of SAM BBOX with gold standard box prompts, and surpasses SOTA on specific datasets such as ISIC and WBC without additional manual prompts. This underscores the important contribution of the GPT-4 to zero-shot segmentation and shows that integrating foundational models such as LLM, VLM, and SAM can increase the ability to address complex issues.

Third, by exploring the selection of bbox prompts generated by VLM, we set confidence score-based selection as a viable method for SAM prompts, noting that increasing the number of alternative boxes could modestly increase the segmentation performance.

We also highlight the limitations of the current method, which are primarily reflected in the zero-shot segmentation performance disparity between radiological and non-radiological images.

The results from our analysis suggests that these factors are not only limitations of the TV-SAM algorithm, but also an inherent issue with all approaches based on the VLM and SAM algorithms. This discrepancy is likely due to these foundational models being trained predominantly on conventional natural images, which are typically obtained through optical imaging captured by cameras. This differs fundamentally in imaging principles and modalities from those of radiological images, resulting in these models' relative inefficiency in processing CT and MR images. Moreover, our study initially explores strategies to choose SAM bbox prompts, providing a general trend for selection but does not specify which alternative box prompt is the optimal choice.

Therefore, future research will focus on further exploring how to increase the zero-shot performance of TV-SAM on radiology images, investigate the best strategy for SAM bbox prompts selection, and will develop a superior fully automatic zero-shot segmentation algorithm.

**Acknowledgment** This work was supported by the grants from National Science and Technology Major Project (Nos. 2021YFF1201200, China), Chinese National Science Foundation (62372316), Sichuan Science and Technology Program (2022YFS0048, 2023YFG0126, and 2024YFHZ0091), the 1·3·5 Project for Disciplines of Excellence, West China Hospital, Sichuan University (Nos. ZYYC21004), and Chongqing Technology Innovation and Application Development Project (CSTB2022TIAD-KPX0067).